\documentclass[letterpaper, 10 pt, journal, twoside]{IEEEtran}

\IEEEoverridecommandlockouts                              

\usepackage{amsmath,amsfonts}

\usepackage{array}
\usepackage{textcomp}
\usepackage{stfloats}
\usepackage{url}
\usepackage{verbatim}
\usepackage{graphicx}
\usepackage{balance}
\usepackage{algorithmic}
\usepackage{adjustbox}
\usepackage{bigstrut}
\usepackage{multirow}
\usepackage{subcaption}
\usepackage{caption} 
\makeatletter
\newcommand\notsotiny{\@setfontsize\notsotiny\@vipt\@viipt}
\makeatother

\usepackage{booktabs}
\usepackage{tabularx}
\usepackage{multirow}
\usepackage[table,xcdraw,dvipsnames]{xcolor}
\definecolor{lightgreen}{rgb}{0.88, 1, 0.88}
\definecolor{lightred}{rgb}{1, 0.88, 0.88}
\definecolor{lightgray}{rgb}{0.83,0.83,0.83}
\usepackage{amssymb}
\usepackage{pifont}

\usepackage{xcolor}

\newcommand{\nBL}{9}
\newcommand{\nBLl}{nine}
\newcommand{\nTr}{9}

\newcommand{\setofBL}{L \in \{1, 10,  20, 30, 40, 60, 80, 120, 240\}}

\newcommand{\urlDataset}{https://github.com/bferrarini/MotionBlurGenerator}
\newcommand{\luzzara}{Luzzara}
\newcommand{\luz}{LZR}
\newcommand{\guastalla}{Guastalla}
\newcommand{\gua}{GST}
\newcommand{\casoni}{Casoni}
\newcommand{\cas}{CSN}

\newcommand{\luzmb}{\luz{}-MBlur}
\newcommand{\guamb}{\gua{}-MBlur}
\newcommand{\casmb}{\cas{}-MBlur}
\newcommand{\luzMX}{\luz{}-Mixed}
\newcommand{\guaMX}{\gua{}-Mixed}
\newcommand{\casMX}{\cas{}-Mixed}

\newcommand{\bferra}{Bruno Ferrarini}
\newcommand{\bferraEM}{bferrarini.ac.uk@gmail.com}

\newcommand{\sheh}{Shoaib Ehsan}

\newcommand{\shehEMs}{s.ehsan@soton.ac.uk}

\newcommand{\mm}{Michael Milford}
\newcommand{\mmEM}{michael.milford@qut.edu.au}

\newcommand{\timur}{Timur Ismagilov}
\newcommand{\timurEM}{ti1e24@soton.ac.uk}

\newcommand{\sdr}{Sarvapali D. Ramchurn}
\newcommand{\sdrEM}{sdr1@soton.ac.uk}

\newcommand{\tvtn}{Tan Viet Tuyen Nguyen}
\newcommand{\tvtnEM}{tuyen.nguyen@soton.ac.uk}

\newcommand{\marklessfootnote}[1]{{\let\thefootnote\relax\footnotetext{#1}}}

\newcommand{\figlabel}{Fig.}

\newcommand{\fig}[1]{\figlabel{} \ref{#1}}

\newcommand{\eq}[1]{Eq. \ref{#1}}

\newcommand{\tab}[1]{Table \ref{#1}}
\newcommand{\sect}[1]{Section \ref{#1}}

\usepackage{enumitem,amssymb}
\usepackage{pifont}
\newlist{todolist}{itemize}{2}
\setlist[todolist]{label=$\square$}
\newcommand{\cmark}{\ding{51}}%
\newcommand{\xmark}{\ding{55}}%

\newcommand{\cmt}[1]{\ignorespaces}

\definecolor{glaucous}{rgb}{0.38, 0.51, 0.71}

\newcommand*{\figtiTFlitefont}{\fontfamily{phv}\selectfont}

\title{
On Motion Blur and Deblurring
in Visual Place Recognition
}

\author{\timur{}$^{1}$*, \bferra{}$^{2}$*, \mm{}$^{3}$ \\ 
\tvtn{}$^{1}$, \sdr{}$^{1}$, \sheh{}$^{1,4}$ 
\thanks{Manuscript received: December 9, 2024; Revised February 18, 2025; Accepted March 13, 2025.}
\thanks{This paper was recommended for publication by Pascal Vasseur upon evaluation of the Associate Editor and Reviewers' comments.
This work was supported by the U.K. Engineering and Physical Sciences Research Council under Grant EP/Y009800/1, Grant EP/V00784X/1 and the EPSRC-UKRI Grant UKRI840 (PRIV-LOC).}
\thanks{*These authors equally contributed to this work.}%
\thanks{$^{1}$\timur{}, \tvtn{}, \sdr{} and \sheh{} are with the School of Electronics and Computer Science, University of Southampton, SO17 1BJ Southampton, U.K. {\tt\footnotesize \timurEM{}, \tvtnEM{}, \sdrEM{}, \shehEMs{}}}%
\thanks{$^{2}$\bferra{} is with MyWay srl, Via Osti, 6 - 29010 Vernasca (PC).  {\tt\footnotesize \bferraEM{}}}%
\thanks{$^{3}$\mm{} is with the QUT Centre for Robotics, School of Electrical
Engineering and Robotics, Brisbane, QLD 4000, Australia. {\tt\footnotesize \mmEM{}}}%
\thanks{$^{4}$\sheh{} is also with the School of Computer Science and Electronic Engineering, University of Essex, Colchester, CO4 3SQ, U.K.}%
\thanks{Digital Object Identifier (DOI): see top of this page.}%
}

\begin{document}

\markboth{IEEE Robotics and Automation Letters. Preprint Version. Accepted March, 2025}
{Ismagilov \MakeLowercase{\textit{et al.}}: On Motion Blur and Deblurring in Visual Place Recognition} 

\setcounter{figure}{-2}

\makeatletter

\let\@oldmaketitle\@maketitle
\renewcommand{\@maketitle}{\@oldmaketitle
\vspace*{1em}
  \begin{center}
    \includegraphics[width=\linewidth]{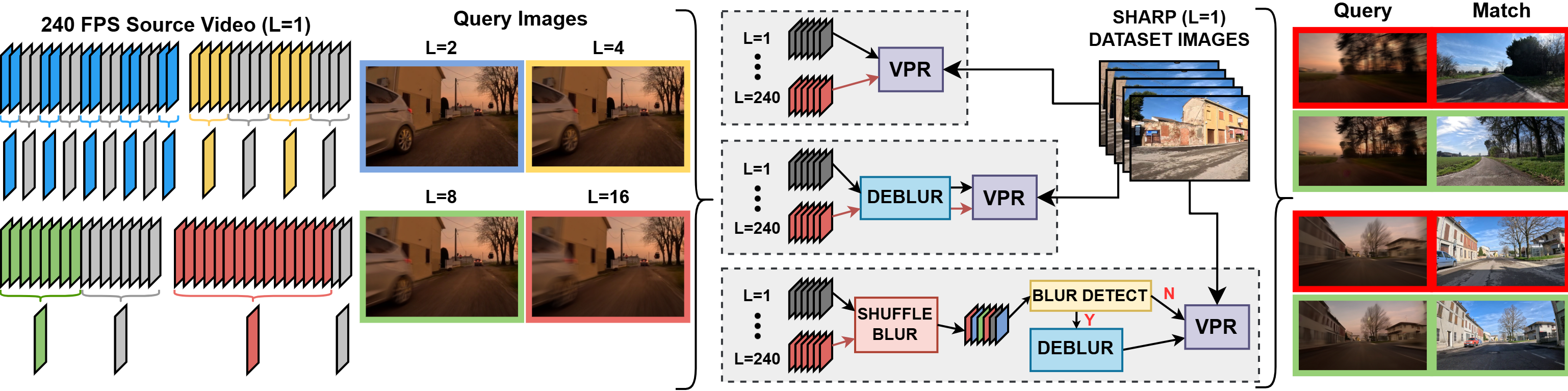}
    \captionof{figure}{We introduce a benchmark featuring motion blur and scene variation and evaluate VPR performance under motion blur, deblurring, and adaptive deblurring. Examples highlight incorrect (red) to correct (green) matches after deblurring.}
    \label{fig:Methods}                         
  \end{center}
    \vspace*{-2em}
    }
\makeatother

\maketitle


\begin{abstract}
Visual Place Recognition (VPR) in mobile robotics enables robots to localize themselves by recognizing previously visited locations using visual data. While the reliability of VPR methods has been extensively studied under conditions such as changes in illumination, season, weather and viewpoint, the impact of motion blur is relatively unexplored despite its relevance not only in rapid motion scenarios but also in low-light conditions where longer exposure times are necessary. Similarly, the role of image deblurring in enhancing VPR performance under motion blur has received limited attention so far. This paper bridges these gaps by introducing a new benchmark designed to evaluate VPR performance under the influence of motion blur and image deblurring. The benchmark includes three datasets that encompass a wide range of motion blur intensities, providing a comprehensive platform for analysis. Experimental results with several well-established VPR and image deblurring methods provide new insights into the effects of motion blur and the potential improvements achieved through deblurring. Building on these findings, the paper proposes adaptive deblurring strategies for VPR, designed to effectively manage motion blur in dynamic, real-world scenarios.
\end{abstract}

%
\section{Introduction}

 \IEEEPARstart{I}{n} mobile robotics, Visual Place Recognition (VPR) allows robots to identify their position by matching visual data to previously encountered locations. It is a challenging task due to potential variations in illumination, weather, season, and viewpoint between the query image and the reference map in real-world scenarios. The impact of these changes on VPR performance has been extensively investigated \cite{7353986,milford2012seqslam,6618963}, through numerous datasets available in the literature \cite{zaffar2020vpr}. 

Motion blur is one of the major challenges remaining for VPR methods. It occurs not only in situations where there is rapid camera motion (such as for fast-flying drones \cite{drones5040121}) but also in the case of relatively slow motion under low-illumination conditions where longer exposure times are necessary, potentially affecting a wide range of VPR applications. Despite this, the impact of motion blur on VPR methods is relatively unexplored. Although there are some datasets available \cite{schops_bad_2019,dai_real-time_2022,mba-vo}, they do come with certain shortcomings (such as limited range of blur intensities, indoor conditions only, low-light conditions only etc.) which make them unsuitable for performing an in-depth investigation. Similarly, to our best knowledge, there has been no work on systematically analyzing the effects of deblurring when used as a mitigation strategy for VPR under a wide variety of blur intensities and in the presence of VPR-specific challenges, such as changes in illumination, weather, and viewpoint. This paper bridges these research gaps and makes three main contributions (as shown in \fig{fig:Methods}): 

\begin{itemize}
  \item 
  We introduce a motion blur-specific benchmark with datasets featuring varying blur levels and VPR-specific appearance changes, enabling comprehensive analysis without field image acquisition. Unlike existing VPR datasets that feature only limited intrinsic blur, our benchmark offers controllable motion blur intensity in outdoor scenes. \fig{fig:samples} shows sample images of the same location in sharp and blurred versions across different traverses. 
  

  \item 
   A comprehensive evaluation of motion blur and deblurring in VPR settings is conducted using the methods depicted in \fig{fig:Methods}. There has been little study on VPR under motion blur, therefore, this analysis leverages the proposed benchmark to empirically examine its impact on the performance of various state-of-the-art VPR techniques.
   
   \item
   Building on the outcomes of the first two contributions, various adaptive deblurring scenarios are evaluated, along with their associated computational costs. These assessments provide valuable insights to guide future experiments involving selective deblurring tailored to VPR-specific challenges, paving the way for advancements in blur-aware VPR techniques.
 \end{itemize}

The paper is organized as follows: \sect{sec:work} reviews related work on VPR under motion blur and deblurring; \sect{sec:building} introduces the Blurry Places benchmark; \sect{sec:setup} details the experimental setup; \sect{sec:results} presents the results; and \sect{sec:conclusions} draws conclusions.

%

\section{Related Work}
\label{sec:work}



This section reviews related work on VPR under motion blur and image deblurring. The impact of motion blur on VPR remains underexplored, with limited and often inadequate datasets available. For example, ETH3D \cite{schops_bad_2019} offers only three sequences of a single scene, while \cite{dai_real-time_2022} matches blurred and sharp images for aerial applications with restricted blur intensity. Similarly, \cite{mba-vo} provides a motion blur-specific dataset but is confined to low-light, indoor, and meter-scale trajectories, making it unsuitable for broader VPR applications like self-driving cars. Most VPR datasets seldom emphasize motion blur (see Table \ref{tab:comparison}) - typically featuring only fixed or incidental blur effects \cite{nyu}, limiting the analysis. Section \ref{sec:building} clarifies that realistic blur via frame averaging requires high frame rate data to capture smooth temporal details for motion interpolation and fine control of blur intensity; however, datasets like Pittsburgh250k \cite{Torii-CVPR2013} are collected at low frame rates, hindering this approach.

\begin{table}[!hb]
    \large
    \centering
    \caption{Dataset comparison: motion blur (MB) and variations in illumination (I), viewpoint (VP), and weather (W).}
    \begin{adjustbox}{width=\columnwidth}
	
\begin{tabular}{@{}l|cccccccc@{}}
    \toprule
    \textbf{Dataset} & \textbf{VPR} & \textbf{Scene} & \textbf{fps} & \textbf{MB Emphasis} & \textbf{I} & \textbf{VP} & \textbf{W} \\
    \midrule
    St Lucia~\cite{Warren2010} 
      & \color{Green}{\cmark}
      & Urban
      & 30Hz
      & \color{red}{\xmark} (Incidental)
      & \color{Green}{\cmark}
      & slight 
      & \color{red}{\xmark} \\
    Oxford RobotCar~\cite{RobotCarDatasetIJRR} 
      & \color{Green}{\cmark}
      & Urban
      & $<$20Hz
      &  \color{red}{\xmark} (Incidental)
      & \color{Green}{\cmark} 
      & \color{red}{\xmark}
      & \color{Green}{\cmark} \\
    Pittsburgh250k~\cite{Torii-CVPR2013} 
      & \color{Green}{\cmark}
      & Urban
      & Low (GSV)
      &  \color{red}{\xmark} (Incidental)
      & \color{Green}{\cmark}
      & \color{Green}{\cmark}
      & \color{red}{\xmark} \\
    KITTI raw~\cite{Geiger2012CVPR} 
      & \color{Green}{\cmark}
      & Urban
      & 10Hz
      & \color{Red}{\xmark} (Incidental)
      & slight 
      & \color{red}{\xmark} 
      & \color{red}{\xmark} \\
    NYU-VPR~\cite{nyu} 
      & \color{Green}{\cmark}
      & Urban
      & -
      & \color{Green}{\cmark} (Incidental)
      & \color{Green}{\cmark} 
      & \color{Green}{\cmark} 
      & \color{Green}{\cmark} \\
    DE-PR~\cite{Wozniak_2021_ICCV} 
      & \color{Green}{\cmark} (Limited)
      & Indoor
      & -
      & \color{Green}{\cmark} (Fixed)
      & \color{red}{\xmark}
      & \color{Green}{\cmark} 
      & \color{red}{\xmark} \\
    ETH3D~\cite{schops_bad_2019} 
      & \color{Green}{\cmark} (Limited)
      & Indoor
      & $\sim$27.1Hz 
      &  \color{Green}{\cmark} (Fixed)
      & \color{Green}{\cmark}
      & \color{Green}{\cmark}
      & \color{red}{\xmark} \\
    MBA-VO~\cite{mba-vo}
      & \color{Green}{\cmark} (Limited)
      & Indoor
      & 27Hz
      & \color{Green}{\cmark} (Fixed)
      & slight 
      & \color{Green}{\cmark}
      & \color{red}{\xmark}\\
    GoPro~\cite{8099518} 
      & \color{red}{\xmark}
      & Variety
      & 240Hz
      & \color{Green}{\cmark}
      & \color{red}{\xmark}
      & \color{Green}{\cmark}
      & \color{red}{\xmark}\\
    HIDE~\cite{shen2020humanawaremotiondeblurring} 
      & \color{red}{\xmark}
      & Variety
      & 240Hz
      & \color{Green}{\cmark} 
      & \color{red}{\xmark}
      & \color{Green}{\cmark}
      & \color{red}{\xmark}\\
    \midrule
    Blurry Places (Ours)
      & \color{Green}{\cmark}
      & Urban, Rural
      & 240Hz
      & \color{Green}{\cmark} (Controlled)
      & \color{Green}{\cmark}
      & \color{Green}{\cmark}
      & \color{Green}{\cmark}\\
    \bottomrule
\end{tabular}
    \end{adjustbox}
    \label{tab:comparison}
\end{table}

Research on deblurring for VPR is also sparse, with existing datasets offering limited blur intensity and diversity. For example, ‘GoPro’ \cite{8099518} uses urban 240fps videos averaging every 7–13 frames, and ‘HIDE’ \cite{shen2020humanawaremotiondeblurring} focuses on outdoor human movement averaging every 11 frames from 240 fps videos, yet neither are structured for VPR. \cite{deblurslam} proposes a SLAM framework that integrates blur detection with DeblurGANv2 deblurring, achieving improved feature matching with direct deblurring, however, they address the method’s limited generalization to diverse scenes. Conversely, \cite{Wozniak_2021_ICCV} introduces a selective deblurring approach for indoor scene classification, showing that detecting and excluding blurred frames improves performance while applying any deblurring does not. Overall, while urban and indoor scene deblurring research exists \cite{Kim_2024_CVPR,articleDLinMD,suin2020spatiallyattentivepatchhierarchicalnetworkadaptive}, there is little analysis of how deblurring affects VPR performance under complex challenges, such as illumination, weather, and viewpoint variations. These gaps highlight the need for more comprehensive studies in this area.

\section{Blurry Places Benchmark}
\label{sec:building}



Blurred frame generation is typically achieved either through blur-kernel-based algorithms or by averaging consecutive, short-exposure frames captured at high frame rates \cite{8099518, wilddeblurring, dynviddeblur, guo_exploring_2021, learntosyn}. The latter method is considered more realistic \cite{guo_exploring_2021}, as it generates spatially varying (non-uniform) blur patterns that closely model the effects caused by perspective shifts and depth variation. In contrast, kernel-based methods often struggle with occlusion, depth variation and deformations, and kernel estimation is computationally expensive and sensitive to noise saturation. Moreover, studies on deblurring models trained on images produced by frame averaging have demonstrated better performance than those trained on synthetically blurred images generated by kernel-based algorithms \cite{wilddeblurring}. Consequently, recent and well-established motion blur datasets \cite{8099518,shen2020humanawaremotiondeblurring,guo_exploring_2021} also employ the frame-averaging approach. 
 
 The blurred images are obtained by averaging the frames of a slow-motion video at 240 fps captured with a GoPro 11 Black. Unlike filter-based methods (e.g., Gaussian filtering), this approach embeds the motion of a video in a blurred image though mimicking the physical image formation in digital cameras \cite{ikeuchi_computer_2021}. In detail, this approach models the image formation as the integral over time of the image illuminating the camera's sensors during exposure time ($\tau$):
\begin{equation}
    \mathbf{I}(x) =  \frac{1}{\tau} \int_{0}^{\tau} \mathbf{I}(t,x(t)) \,dt\,. 
    \label{eq:image_formation}
\end{equation}
$\mathbf{I}(t,x(t))$ is the image to which the sensor is exposed at the time $t$ during the exposure time ($\tau$). $\mathbf{I}(x)$ is a frame captured by the camera.
This model describes how $\mathbf{I}(t,x(t))$ variations over time induce motion blur. 
Faster movements generate wider variations and, therefore, more intense motion blur in the part of the sensor where they occur. 

We build our datasets from a discrete sequence of video frames, thus \eq{eq:image_formation} is approximated with the following series:

\begin{equation}
    \mathbf{I}(x) \approx \frac{1}{n} \sum_{i=0}^{n-1} \mathbf{I_i}(x) \,. 
    \label{eq:image_formation_discrete}
\end{equation}

The integral is replaced by a sum of $n$ sharp frames indicated with $\mathbf{I_i}(x)$. 
\eq{eq:image_formation_discrete} is then parameterized as follows to build multiple blurring intensities:
\begin{equation}
    \mathbf{B_j^L}(x) = \frac{1}{L} \sum_{i=j}^{j+L} \mathbf{I_i}(x) \,. 
    \label{eq:image_formation_syntetic}
\end{equation}

 \begin{figure}[!t]
\centering
\vspace*{1ex}
\includegraphics[width=0.85\columnwidth{}]{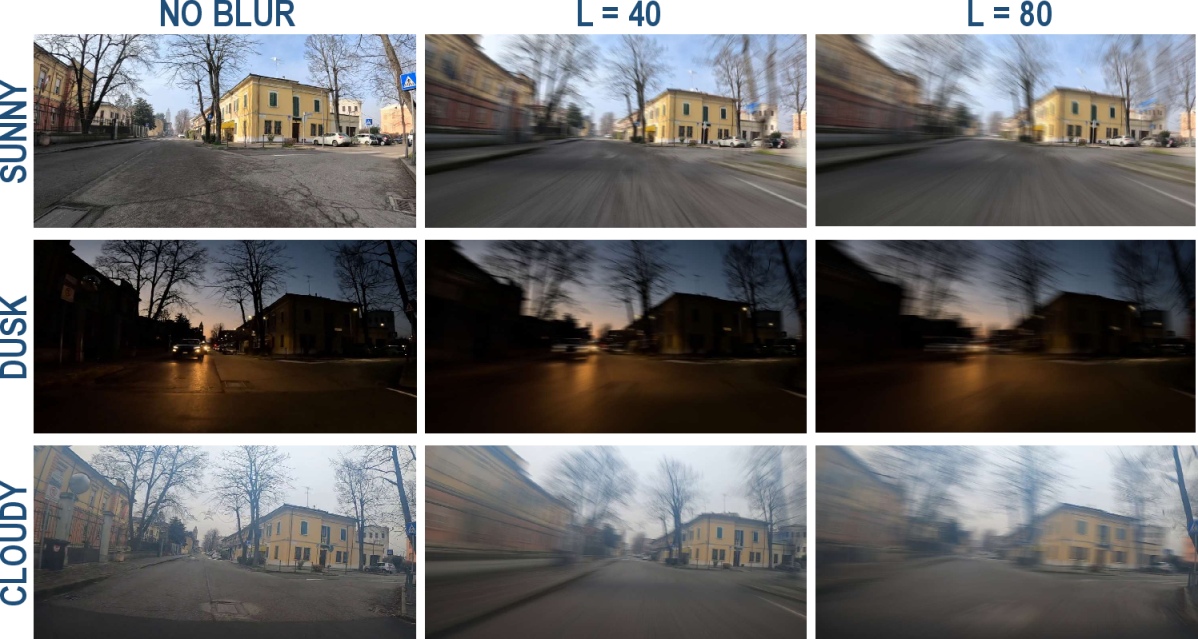}
\caption{A place in \luzzara{} in 3 traverses and blur intensities.}
\vspace*{-1em}
\label{fig:samples}
\end{figure}

$\mathbf{B_j^L}$ is the blurred image from averaging $L$ sharp frames starting at frame $j$ of the source video, with increasing blur intensity as $L$ increases. We use $L$ in the remainder of the paper to indicate the amount of motion blur of an image, with $L = 1$ corresponding to a sharp image.
\fig{fig:Methods} illustrates the synthetic blur generation process from \eq{eq:image_formation_syntetic} with sample images at several blur levels; the colour code used serves only to highlight the contributing sharp frames.


Sample images with motion embedding are shown in \fig{fig:samples}, where some areas remain relatively sharp depending on movement direction.
Note that $L$ relates to the exposure time in \eq{eq:image_formation} via the source video frame rate ($V_{fps}$).

\begin{equation}
    \tau \approx \hat{\tau} = \frac{L}{V_{fps}}.
    \label{eq:virtual_tau}
\end{equation}

%
Here, $\hat{\tau}$ approximates the shutter time $\tau$ of the blur model described by \eq{eq:image_formation}. The higher the video frame rate, $V_{fps}$, the finer the control on the blur level, as $L$ can vary in a wider range within the same exposure time. For example, consider two videos: one collected at 240 fps and the other at 30 fps at the same speed. With $L=2$, the 240 fps video results in a 120 fps blurred dataset, while the 30 fps video only achieves a 15 fps blurred dataset with significantly higher exposure time. Alternatively, using the exposure time $\hat{\tau}$ as a reference, the same blur can be applied to videos at different $V_{fps}$. For example, the blurring corresponding to $\hat{\tau} = 0.5$s requires $L = 120$ and $L = 15$ for the 240 and 30 fps videos, respectively. Furthermore, a higher frame rate enhances motion continuity at the same speed, producing a more realistic motion blur through averaging.
Driven by this consideration, the raw data to build the proposed dataset are videos captured with a GoPro 11 black at the highest possible frame rate of 240 fps.
The benchmark is organized in datasets. A dataset includes several sets of images captured along the same route at different times, so it is always possible to use one as a reference (previously visited locations) and the other as a query (a second traverse through the route). The traverses are generally different due to driving trajectories causing lateral shifts, weather, illumination, seasons, and dynamic elements like pedestrians, appearing only in one of the traverses. These appearance changes are used jointly with motion blur to set up different testing scenarios. 
Specifically, the proposed benchmarks include \nTr{} traverses captured along the three outdoor routes in urban and country-side environments. These are named as the small towns where they are recorded: \luzzara{} (\luz{}), \guastalla{} (\gua{}) and \casoni{} (\cas{}), in Italy. The recording list is shown in \tab{tab:record-list}, indicating the environmental conditions during the recording, the duration of the video, and the number of sharp frames it contains.  

Every video in \tab{tab:record-list} was processed with the method described by \eq{eq:image_formation_syntetic} using the following \nBLl{} values of $L$:

\begin{equation}
   \setofBL{}\,,
    \label{eq:BL}
\end{equation}
\noindent
where $L=1$ (no blur) establishes a baseline for evaluating the impact of blurring. We found empirically that this set of blur levels ensures a comprehensive characterization of VPR performance. Nevertheless, any blur level can be generated using the provided scripts and data\footnote{\urlDataset{}\label{blurgen}}, including extending datasets captured in the same manner to domains beyond those tested (e.g., extreme weather, indoors).
Traversals can be combined into query-reference pairs to conduct experiments on motion blur at any intensity in $L$, with or without the environmental conditions available (e.g. night). Moreover, selecting the datasets with $L=1$ (no blur), the resulting benchmark can be used to assess VPR under the viewing conditions as with any other well-established VPR datasets such as St. Lucia \cite{Warren2010} and Oxford Car \cite{RobotCarDatasetIJRR}. 



%
\begin{table}[t]
    \centering
    \caption{Traverse recorded in \luzzara{} (\luz{}), \guastalla{} (\gua{}), \casoni{} (\cas{}) routes.}
    \begin{adjustbox}{width=\columnwidth}
	    \begin{tabular}{|r|cccccc|}
    \multicolumn{1}{c}{\textbf{Track}} & \textbf{Traverse} & \textbf{Time} & \textbf{Condition} & \textbf{length} & \textbf{duration} & \multicolumn{1}{c}{\textbf{Frames}} \bigstrut[b]\\
    \hline
    \hline
    \multicolumn{1}{|l|}{\luzzara{}} & 01    & Morning & Cloudy & \multirow{3}[2]{*}{3.2 Km} & 06:52 & 98810 \bigstrut[t]\\
    \multicolumn{1}{|c|}{(\luz{})} & 02    & Dusk  & Sunny &       & 06:33 & 94442 \\
          & 03    & Morning & Sunny &       & 05:57 & 85679 \bigstrut[b]\\
    \hline
    \multicolumn{1}{|l|}{\guastalla{}} & 01    & Afternoon & Sunny & \multirow{3}[2]{*}{3.6 Km} & 05:59 & 86129 \bigstrut[t]\\
    \multicolumn{1}{|c|}{(\gua{})} & 02    & Afternoon & Sunny &       & 05:38 & 81164 \\
          & 03    & Morning & Sunny &       & 05:47 & 83272 \bigstrut[b]\\
    \hline
    \multicolumn{1}{|l|}{\casoni{}} & 01    & Noon  & Cloudy & \multirow{3}[2]{*}{4.3 Km} & 06:08 & 77889 \bigstrut[t]\\
    \multicolumn{1}{|c|}{(\cas{})} & 02    & Dusk  & Sunny &       & 05:24 & 88476 \\
          & 03    & Morning & Sunny &       & 05:28 & 78131 \bigstrut[b]\\
    \hline
    \end{tabular}%
    \end{adjustbox}
    \label{tab:record-list}
    \vspace*{-1em}
\end{table}
\section{Experimental Setup}
\label{sec:setup}
\begin{figure*}[!hbp]
\centering
\input{include/COMBINED_AUC_PLOTS}
\vspace{-0.5cm}
\caption{AUC at various motion blur levels. Blur-only traverse results are shown at the top, while the bottom row shows the results for mixed conditions. The bars indicate the 95\% confidence interval obtained using block bootstrapping \cite{blockboot}.}
	\label{fig:performance}
\vspace*{-1em}
\end{figure*}

To demonstrate the utilization of the proposed benchmark, we use the datasets formed by the pairs enlisted in \tab{tab:datasets}. The first three rows use the same traverse as both query and reference. All the experiments use sharp images as a reference, while those in the query traverse are blurred at \nBL{} increasing intensities as in \eq{eq:BL}. This setup excludes any other change but motion blur from the VPR performance analysis. The last three rows present more challenging (and realistic) scenarios where the query and reference images are captured in different traverses. These scenarios include $L=1$ to establish a performance baseline under weather, illumination, and viewpoint, along with additional motion blur at increasing levels of $L$. The same traverses are used for analysis of direct deblurring, and a subset of them for adaptive deblurring, where the blur intensities are shuffled into one sequence as described in their respective results.

\begin{table}[!t]
    \centering
    \caption{The traverse pairs used for the experiments. Each comes with one or more appearance changes: Motion Blur (\textbf{MB}), Weather (\textbf{W}), Illumination (\textbf{I}), and ViewPoint (\textbf{VP}).}
    \begin{adjustbox}{width=\columnwidth}
    \begin{tabular}{|c|cc|cccc|}
    \multicolumn{1}{c}{\textbf{Pair ID}} & \textbf{Query} & \multicolumn{1}{c}{\textbf{Reference}} & \textbf{MB} & \textbf{W} & \textbf{I} & \multicolumn{1}{c}{\textbf{VP}} \bigstrut[b]\\
    \hline
    \hline
    \multirow{2}[2]{*}{\luzmb{}} & \multicolumn{2}{c|}{01-Evening-Sunny} & \multirow{2}[2]{*}{X} & \multirow{2}[2]{*}{} & \multirow{2}[2]{*}{} & \multirow{2}[2]{*}{} \bigstrut[t]\\
          & \multicolumn{2}{c|}{(411 images)} &       &       &       &  \bigstrut[b]\\
    \hline
    \multirow{2}[2]{*}{\guamb{}} & \multicolumn{2}{c|}{01-Afternoon-Sunny} & \multirow{2}[2]{*}{X} & \multirow{2}[2]{*}{} & \multirow{2}[2]{*}{} & \multirow{2}[2]{*}{} \bigstrut[t]\\
          & \multicolumn{2}{c|}{(358 images)} &       &       &       &  \bigstrut[b]\\
    \hline
    \multirow{2}[2]{*}{\casmb{}} & \multicolumn{2}{c|}{01-Noon-Cloudy} & \multirow{2}[2]{*}{X} & \multirow{2}[2]{*}{} & \multirow{2}[2]{*}{} & \multirow{2}[2]{*}{} \bigstrut[t]\\
          & \multicolumn{2}{c|}{(324 images)} &       &       &       &  \bigstrut[b]\\
    \hline
    \multirow{2}[2]{*}{\scriptsize{\luzMX{}}} & 02-Morning-Sunny & 03-Morning-Cloudy & \multirow{2}[2]{*}{X} & \multirow{2}[2]{*}{X} & \multirow{2}[2]{*}{} & \multirow{2}[2]{*}{X} \bigstrut[t]\\
          & (393 images) & (356 images) &       &       &       &  \bigstrut[b]\\
    \hline
    \multirow{2}[2]{*}{\scriptsize{\guaMX{}}} & \multicolumn{1}{l}{02-Afternoon-Sunny} & \multicolumn{1}{l|}{01-Afternoon-Sunny} & \multirow{2}[2]{*}{X} & \multirow{2}[2]{*}{} & \multirow{2}[2]{*}{} & \multirow{2}[2]{*}{X} \bigstrut[t]\\
          & (338 images) & (358 images) &       &       &       &  \bigstrut[b]\\
    \hline
    \multirow{2}[2]{*}{\scriptsize{\casMX{}}} & 02-Dusk-Sunny & \multicolumn{1}{l|}{03-Morning-Sunny} & \multirow{2}[2]{*}{X} & \multirow{2}[2]{*}{} & \multirow{2}[2]{*}{X} & \multirow{2}[2]{*}{X} \bigstrut[t]\\
          & (325 images) & (325 images) &       &       &       &  \bigstrut[b]\\
    \hline
    \end{tabular}%

    \end{adjustbox}
    \label{tab:datasets}
    \vspace*{-1em}
\end{table}

Several well-established VPR methods were systematically evaluated in our benchmark to cover several approaches, including MixVPR \cite{ali-bey_mixvpr_2023}, AnyLoc \cite{keetha_anyloc_2023}, CosPlace \cite{Berton_CVPR_2022_CosPlace}, EigenPlaces \cite{Berton_2023_EigenPlaces}, HDC-DELF \cite{neubert_hyperdimensional_2021}, FloppyNet \cite{floppynet}, Patch-NetVLAD \cite{hauslerPatchNetvlad2021}, ORB \cite{rublee2011orb}, and SAD \cite{milford2012seqslam}, with blur intensities ranging from L=1 to L=240. Patch-NetVLAD adopts a two-stage retrieval process with its own matching step, while the others use a single-stage approach with cosine similarity for descriptor comparison. Offline deblurring is applied to each level of blur intensity in all datasets, using three state-of-the-art deblurring methods: DeblurGANv2 \cite{Kupyn_2019_ICCV}, FFTFormer \cite{kong2022efficientfrequencydomainbasedtransformers} and GShift-Net \cite{Li_2023_CVPR}. We then perform the same VPR analysis to get a clear comparison of the impacts of deblurring across blur and scene variation. We evaluate VPR performance using the Area Under Curve (AUC) of Precision-Recall curves. A higher AUC indicates stronger performance, reflecting consistent retrieval of relevant images with minimal false positives, whereas a lower AUC implies difficulty maintaining high precision and recall simultaneously. For VPR methods, we follow the setup shared by the authors, while deblurring is applied to the raw benchmark images to ensure an unbiased evaluation of VPR performance. Adaptive deblurring employs Laplacian variance for blur detection due to its simplicity and efficiency \cite{drones5040121}. Empirical tests revealed a wide variance gap between sharp and severely blurred images (L$>$40), and results confirm deblurring is only significant with frequent, high blur intensities. Consequently, fixed thresholds of 50 for the shuffled \casMX{} dataset and 200 for the shuffled \guaMX{} and \luzMX{} datasets achieved near-perfect classification of such blurred frames.

\section{Results and Discussion}
\label{sec:results}

\subsection{Motion Blur}
The effect of motion blur is examined using the traverse pairs \luzmb{}, \guamb{}, and \casmb{}. These three datasets use the same traverse as both reference and query, so the images differ only by the amount of motion blur. The results are shown in \fig{fig:performance} for increasing blurring intensities in query images. All the plots start from $\text{AUC} = 1$ as at $L = 1$, query and reference images are identical.
As expected, every VPR method is negatively impacted. FloppyNet achieves the most robust performance regardless of it being a Binary Neural Network (BNN) - a highly compact variant of neural network where the weights and activations are constrained to 1-bit values to reduce memory and computational cost. FloppyNet's shallow architecture of just three layers makes it tolerant to viewpoint-free appearance changes \cite{floppynet} and exceeds severe motion blur performance in \guamb{} and \casmb{}. SAD is considered to tolerate viewpoint-free changes \cite{sunderhauf2013we}, however, it exhibits steeper performance decay, dropping quickly to an AUC between 0.2 and 0.3. This decay is greater with \luzmb{}, which may arise from the dataset's reduced lighting and contrast. Patch-NetVLAD starts strongly but degrades quickly, achieving a lower performance among the learning-based methods for larger blur, and comparable or even worse performance than ORB-VLAD in \luzmb{} and \guamb{}. ORB-VLAD, being non-learning-based, performs competitively with the learning-based methods on \luzmb{} and \casmb{}, however quickly suffers on \guamb{}. The remaining models exhibit similar performance and degrading behaviour, being the more robust choices out of all the methods capable of handling large motion blur reasonably well. This is reflected in the measured confidence ranges, with top-performing models having little variance in performance, but becoming less certain as blur intensity increases. We expect AnyLoc to outperform in diverse scenes as it utilizes general-purpose features from large-scale pretrained models. Overall, \casmb{} proves to be the variation-free dataset that VPR models struggled the most on with severe blur intensity.
%
%

%
\begin{table*}[!hbp]
\centering
\renewcommand{\arraystretch}{1.2}
\setlength{\tabcolsep}{3pt}
\notsotiny
\caption{AUC of VPR methods with different deblurring methods on \casMX{}. Bold indicates the best performing combinations within each query blur level. Improvements after deblurring highlighted in green; deterioration in red. }
\begin{tabularx}{0.9\textwidth}{c c *{9}{|>{\centering\arraybackslash}X}| c c|>{\centering\arraybackslash}p{2cm}} 
\multicolumn{2}{c}{} & \multicolumn{10}{c}{\textbf{Query Blur Level}} \\ 
\hline
\textbf{VPR Model} & \textbf{Deblur Method} & \textbf{001} & \textbf{010} & \textbf{020} & \textbf{030} & \textbf{040} & \textbf{060} & \textbf{080} & \textbf{120} & \textbf{240} & Avg. & Std. & Avg. Deblur + Extract time / query (ms) \\
\hline
MixVPR & No Deblur & 0.98 & 0.97 & 0.96 & 0.94 & 0.92 & 0.85 & 0.79 & 0.70 & 0.56 & 0.85 & 0.13 & 3.72 \\
& DeblurGANv2 & \cellcolor{lightgray}0.98 & \cellcolor{lightgray}0.97 & \cellcolor{lightgreen}\underline{\textbf{0.97}} & \cellcolor{lightgreen}\underline{\textbf{0.97}} & \cellcolor{lightgreen}\underline{\textbf{0.95}}& \cellcolor{lightgreen}\underline{\textbf{0.91}} & \cellcolor{lightgreen}\underline{\textbf{0.88}}& \cellcolor{lightgreen}\underline{\textbf{0.81}} & \cellcolor{lightgreen}0.59 & \underline{\textbf{0.89}} & \underline{\textbf{0.11}} & 38.41  \\
& GShift-Net & \cellcolor{lightgray}0.98 & \cellcolor{lightgreen}\underline{\textbf{0.98}} & \cellcolor{lightgreen}\underline{\textbf{0.97}} & \cellcolor{lightgreen}0.95 & \cellcolor{lightgreen}0.93 & \cellcolor{lightgreen}0.88 & \cellcolor{lightgreen}0.82 &
\cellcolor{lightgreen}0.74 & \cellcolor{lightgreen}\underline{\textbf{0.65}} & 0.88 & \underline{\textbf{0.11}} & 394.65\\
& FFTFormer & \cellcolor{lightgray} 0.98 & \cellcolor{lightgray}0.97 & \cellcolor{lightgray}0.96 & \cellcolor{lightgreen}0.95 & \cellcolor{lightgreen}0.93 & \cellcolor{lightgreen}0.88 & \cellcolor{lightgreen}0.84 & \cellcolor{lightgreen}0.75 & \cellcolor{lightred}0.52 & 0.86 & 0.14 & 590.33 \\
\hline

AnyLoc & No Deblur & 0.93 & 0.91 & 0.89 & 0.86 & 0.80 & 0.74 & 0.69 & 0.60 & 0.48 & 0.76 & \underline{\textbf{0.14}} & 624.84 \\
    & DeblurGANv2 & \cellcolor{lightgray}0.93 & \cellcolor{lightgreen}0.92 & \cellcolor{lightgreen}\underline{\textbf{0.91}} & \cellcolor{lightgreen}0.88 & 
    \cellcolor{lightgreen}\underline{\textbf{0.83}} & \cellcolor{lightgreen}0.75 &
    \cellcolor{lightgreen}\underline{\textbf{0.75}} &\cellcolor{lightgreen}0.63 &\cellcolor{lightred}0.43 & \underline{\textbf{0.78}} & 0.15 & 659.53 \\
& GShift-Net & \cellcolor{lightgreen}\underline{\textbf{0.94}} & \cellcolor{lightgreen}\underline{\textbf{0.93}} & \cellcolor{lightgreen}0.90 & \cellcolor{lightred}0.85 & \cellcolor{lightgreen}\underline{\textbf{0.83}} & \cellcolor{lightred}0.72 & \cellcolor{lightgreen}0.72 & \cellcolor{lightgreen}\underline{\textbf{0.64}} & \cellcolor{lightgreen}\underline{\textbf{0.49}} & \underline{\textbf{0.78}} & \underline{\textbf{0.14}} & 1015.77\\

& FFTFormer & \cellcolor{lightgreen}\underline{\textbf{0.94}} & \cellcolor{lightgreen}0.92 & \cellcolor{lightgreen}0.90 & \cellcolor{lightgreen}\underline{\textbf{0.89}} & \cellcolor{lightgreen}0.82 & \cellcolor{lightgreen}\underline{\textbf{0.76}} & \cellcolor{lightgreen}0.72 & \cellcolor{lightgreen}0.63 & \cellcolor{lightred}0.44 & \underline{\textbf{0.78}} & 0.15 & 1211.45\\
\hline

CosPlace & No Deblur & 0.95 & 0.96 & 0.94 & 0.93 & 0.92 & 0.87 & 0.82 & 0.70 & 0.59 & 0.85 & 0.12 & 3.78 \\
& DeblurGANv2 & \cellcolor{lightgreen}0.96 & \cellcolor{lightgreen}\underline{\textbf{0.97}} & \cellcolor{lightgreen}\underline{\textbf{0.97}} & \cellcolor{lightgreen}\underline{\textbf{0.96}} &
\cellcolor{lightgreen}\underline{\textbf{0.96}} &
\cellcolor{lightgreen}\underline{\textbf{0.95}} & \cellcolor{lightgreen}\underline{\textbf{0.91}} & \cellcolor{lightgreen}\underline{\textbf{0.87}} & \cellcolor{lightgreen}0.66 & \underline{\textbf{0.91}} & \underline{\textbf{0.09}} & 38.47\\
& GShift-Net & \cellcolor{lightgreen}\underline{\textbf{0.98}} & \cellcolor{lightgreen}\underline{\textbf{0.97}} & \cellcolor{lightgreen}\underline{\textbf{0.97}} & \cellcolor{lightgreen}0.94 & \cellcolor{lightgray}0.92 & \cellcolor{lightgreen}0.89 & \cellcolor{lightgreen}0.84 & \cellcolor{lightgreen}0.71 & \cellcolor{lightgreen}\underline{\textbf{0.74}} & 0.89 & \underline{\textbf{0.09}} & 394.71 \\
& FFTFormer & \cellcolor{lightgray}0.95 & \cellcolor{lightred}0.95 & \cellcolor{lightgray}0.94 & \cellcolor{lightgray}0.93 & \cellcolor{lightred}0.91 & \cellcolor{lightgreen}0.88 & \cellcolor{lightgreen}0.84 & \cellcolor{lightgreen}0.79 & \cellcolor{lightgreen}0.62 & 0.87 & 0.10 & 590.39\\
\hline

EigenPlaces & No Deblur & 0.97 & 0.95 & 0.94 & 0.93 & 0.91 & 0.89 & 0.84 & 0.76 & 0.56 & 0.86 & 0.12 & 3.76\\
& DeblurGANv2 & \cellcolor{lightgray}0.97 & \cellcolor{lightgreen}0.97 & \cellcolor{lightgreen}\underline{\textbf{0.96}} & \cellcolor{lightgreen}\underline{\textbf{0.96}} & \cellcolor{lightgreen}\underline{\textbf{0.95}} & \cellcolor{lightgreen}\underline{\textbf{0.93}} & \cellcolor{lightgreen}\underline{\textbf{0.88}} & \cellcolor{lightgreen}\underline{\textbf{0.84}} & \cellcolor{lightgreen}0.72 & \underline{\textbf{0.91}} & \underline{\textbf{0.08}} & 38.45\\
& GShift-Net & \cellcolor{lightgreen}\underline{\textbf{0.98}} & \cellcolor{lightgreen}\underline{\textbf{0.98}} & \cellcolor{lightgreen}\underline{\textbf{0.96}} & \cellcolor{lightgreen}0.95 & \cellcolor{lightgreen}0.92 & \cellcolor{lightgray}0.89 & \cellcolor{lightgray}0.84 & \cellcolor{lightgreen}0.77 & \cellcolor{lightgreen}\underline{\textbf{0.75}} & 0.89 & \underline{\textbf{0.08}} & 394.69\\
& FFTFormer & \cellcolor{lightred}0.96 & \cellcolor{lightgray}0.95 & \cellcolor{lightgray}0.94 & \cellcolor{lightgray}0.93 & \cellcolor{lightgreen}0.92 & \cellcolor{lightgreen}0.90 & \cellcolor{lightgreen}0.86 & \cellcolor{lightgreen}0.79 & \cellcolor{lightgreen}0.62 & 0.87 & 0.10 & 590.37\\
\hline

HDCDELF & No Deblur & 0.93 & 0.91 & 0.88 & 0.84 & 0.82 & 0.76 & 0.73 & 0.63 & 0.44 & 0.77 & 0.14 & 610.38\\
    & DeblurGANv2 & \cellcolor{lightgreen}0.94 & \cellcolor{lightgreen}\underline{\textbf{0.93}} & \cellcolor{lightgreen}\underline{\textbf{0.92}} & \cellcolor{lightgreen}\underline{\textbf{0.90}} & \cellcolor{lightgreen}\underline{\textbf{0.86}} & \cellcolor{lightgreen}\underline{\textbf{0.83}} & \cellcolor{lightgreen}\underline{\textbf{0.78}} & \cellcolor{lightgreen}\underline{\textbf{0.74}} & \cellcolor{lightgreen}0.53 & \underline{\textbf{0.82}} & \underline{\textbf{0.12}} & 645.07 \\
& GShift-Net & \cellcolor{lightgreen}\underline{\textbf{0.95}} & \cellcolor{lightgreen}0.92 & \cellcolor{lightgray}0.88 & \cellcolor{lightred}0.83 & \cellcolor{lightred}0.78 & \cellcolor{lightred}0.74 & \cellcolor{lightred}0.70 & \cellcolor{lightred}0.62 & \cellcolor{lightgreen}\underline{\textbf{0.57}} & 0.78 & \underline{\textbf{0.12}} & 1001.31 \\
& FFTFormer & \cellcolor{lightgreen}0.94 & \cellcolor{lightgreen}\underline{\textbf{0.93}} & \cellcolor{lightgreen}0.90 & \cellcolor{lightgreen}0.87 & \cellcolor{lightgreen}0.84 & \cellcolor{lightgreen}0.79 & \cellcolor{lightgreen}0.75 & \cellcolor{lightgreen}0.68 & \cellcolor{lightgreen}0.49 & 0.80 & 0.13 & 1196.99\\
\hline

P-NetVLAD & No Deblur & \underline{\textbf{0.88}} & \underline{\textbf{0.81}} & 0.76 & 0.71 & 0.67 & 0.61 & 0.46 & 0.31 & 0.15 & 0.59 & 0.22 & 10.49\\
    & DeblurGANv2 & \cellcolor{lightred}0.86 & \cellcolor{lightred}0.78 & \cellcolor{lightgreen}\underline{\textbf{0.81}} & \cellcolor{lightgreen}\underline{\textbf{0.77}} & \cellcolor{lightgreen}\underline{\textbf{0.75}} & \cellcolor{lightgreen}\underline{\textbf{0.69}} & \cellcolor{lightgreen}\underline{\textbf{0.61}} & \cellcolor{lightgreen}\underline{\textbf{0.51}} & \cellcolor{lightgreen}0.24 & \underline{\textbf{0.66}} & 0.18 & 45.18 \\
& GShift-Net &\cellcolor{lightred}0.78 & \cellcolor{lightred}0.79 & \cellcolor{lightred}0.74 & \cellcolor{lightred}0.68 & \cellcolor{lightred}0.65 & \cellcolor{lightred}0.50 & \cellcolor{lightgreen}0.49 & \cellcolor{lightgreen}0.33 & \cellcolor{lightgreen}\underline{\textbf{0.30}} & 0.58 & \underline{\textbf{0.17}} & 401.42 \\
& FFTFormer & \cellcolor{lightred}0.82 & \cellcolor{lightred}0.80 & \cellcolor{lightgreen}0.79 & \cellcolor{lightgreen}\underline{\textbf{0.77}} & \cellcolor{lightgreen}0.68 & \cellcolor{lightgreen}0.63 & \cellcolor{lightgreen}0.52 & \cellcolor{lightgreen}0.39 & \cellcolor{lightgray}0.15 & 0.61 & 0.21 & 597.10\\
\hline

FloppyNet & No Deblur & 0.86 & 0.81 & 0.74 & 0.68 & 0.64 & 0.55 & 0.50 & 0.37 & 0.28 & 0.60 & 0.18 & 41.63\\
    & DeblurGANv2 & \cellcolor{lightgray}0.86 & \cellcolor{lightgreen}\underline{\textbf{0.84}} & \cellcolor{lightgreen}\underline{\textbf{0.80}} & \cellcolor{lightgreen}\underline{\textbf{0.78}} & \cellcolor{lightgreen}\underline{\textbf{0.75}} & \cellcolor{lightgreen}\underline{\textbf{0.70}} & \cellcolor{lightgreen}\underline{\textbf{0.59}} & \cellcolor{lightgreen}\underline{\textbf{0.48}} & \cellcolor{lightgreen}0.34 & \underline{\textbf{0.68}} & \underline{\textbf{0.16}} & 76.32\\
& GShift-Net & \cellcolor{lightgray}0.86 & \cellcolor{lightgreen}0.82 & \cellcolor{lightgreen}0.75 & \cellcolor{lightgreen}0.69 & \cellcolor{lightgray}0.64 & \cellcolor{lightgreen}0.57 & \cellcolor{lightgreen}0.52 & \cellcolor{lightgreen}0.41 & \cellcolor{lightgreen}\underline{\textbf{0.34}} & 0.62 & \underline{\textbf{0.16}} & 432.56 \\
& FFTFormer & \cellcolor{lightgray}0.86 & \cellcolor{lightgreen}0.82 & \cellcolor{lightgreen}0.75 & \cellcolor{lightgreen}0.70 & \cellcolor{lightgreen}0.66 & \cellcolor{lightgreen}0.57 & \cellcolor{lightgreen}0.51 & \cellcolor{lightgreen}0.40 & \cellcolor{lightgreen}0.29 & 0.62 & 0.18 & 628.24 \\
\hline

ORB-VLAD & No Deblur & \underline{\textbf{0.23}} & 0.16 & 0.10 & 0.05 & 0.05 & 0.04 & 0.04 &\underline{\textbf{ 0.21}} & 0.02 & 0.10 & 0.07 & 17.26\\
    & DeblurGANv2 & \cellcolor{lightred}0.18 & \cellcolor{lightgreen}\underline{\textbf{0.19}} & \cellcolor{lightgreen}0.14 & \cellcolor{lightgreen}0.09 & \cellcolor{lightgreen}\underline{\textbf{0.08}} & \cellcolor{lightgreen}\underline{\textbf{0.07}} & \cellcolor{lightred}0.03 & \cellcolor{lightred}0.04 & \cellcolor{lightred}0.01 & 0.09 & 0.06 & 51.95 \\
& GShift-Net & \cellcolor{lightred}0.22 & \cellcolor{lightgray}0.16 & \cellcolor{lightgreen}\underline{\textbf{0.16}} & \cellcolor{lightgreen}\underline{\textbf{0.12}} & \cellcolor{lightgreen}\underline{\textbf{0.08}} & \cellcolor{lightred}0.03 & \cellcolor{lightgreen}\underline{\textbf{0.05}} & \cellcolor{lightred}0.03 & \cellcolor{lightgreen}\underline{\textbf{0.04}} & 0.10 & 0.06 & 408.19\\
& FFTFormer & \cellcolor{lightred}0.21 & \cellcolor{lightred}0.13 & \cellcolor{lightgray}0.10 & \cellcolor{lightgreen}0.07 & \cellcolor{lightgreen}\underline{\textbf{0.08}} & \cellcolor{lightgreen}0.06 & \cellcolor{lightgray}0.04 & \cellcolor{lightred}0.01 & \cellcolor{lightred}0.01 & 0.08 & 0.06 & 603.87\\
\hline

SAD & No Deblur & 0.04 & 0.04 & 0.02 & 0.03 & 0.03 & 0.01 & 0.02 & 0.03 & 0.03 & 0.03 & 0.01 & 3.99\\
    & DeblurGANv2 & \cellcolor{lightgray}0.04 & \cellcolor{lightred}0.03 & \cellcolor{lightgreen}\underline{\textbf{0.04}} & \cellcolor{lightgray}0.03 & \cellcolor{lightgreen}\underline{\textbf{0.04}} & \cellcolor{lightgreen}\underline{\textbf{0.02}} & \cellcolor{lightgray}0.02 & \cellcolor{lightred}0.02 & \cellcolor{lightgray}0.03 & 0.03 & 0.01 & 38.68\\
& GShift-Net & \cellcolor{lightgray}0.04 & \cellcolor{lightred}0.03 & \cellcolor{lightgray}0.02 & \cellcolor{lightred}0.02 & \cellcolor{lightgray}0.03 & \cellcolor{lightgreen}\underline{\textbf{0.02}} & \cellcolor{lightgray}0.02 & \cellcolor{lightgreen}\underline{\textbf{0.04}} & \cellcolor{lightgreen}\underline{\textbf{0.04}} & 0.03 & 0.01 & 394.92\\
& FFTFormer & \cellcolor{lightred}0.03 & \cellcolor{lightgreen}\underline{\textbf{0.05}} & \cellcolor{lightgray}0.02 & \cellcolor{lightgreen}\underline{\textbf{0.04}} & \cellcolor{lightred}0.02 & \cellcolor{lightgreen}\underline{\textbf{0.02}} & \cellcolor{lightgreen}\underline{\textbf{0.03}} & \cellcolor{lightred}0.01 & \cellcolor{lightred}0.02 & 0.03 & 0.01 & 590.60\\
\hline
\end{tabularx}
\label{tab:deblurred_results}
\vspace*{-1em}
\end{table*}
\subsection{Mixed Conditions} 
The mixed-condition datasets used for these experiments are shown in the last three rows of \tab{tab:datasets}, combining motion blur with other appearance variations that are typical of VPR applications, such as viewpoint, weather, and illumination. \fig{fig:performance} shows the AUC trend for increasing motion blur, with $L = 1$ representing the baseline of VPR performance, which degrades with
the addition of motion blur. Regardless of the model, \guaMX{} proves to be a particularly challenging dataset for VPR, indicating that this query traverse introduces non-trivial examples that shift AUC down. The difficulty of these traverses is highlighted by the wider width of the 95\% confidence intervals compared to the intervals in the blur-only traverses. AnyLoc extracts per-pixel features using large-scale pretrained models, making it better suited for general, broad-spectrum applications. With increasing motion blur, AnyLoc performs with little degradation on the \luzMX{} and \guaMX{} datasets, highlighting its robustness to blur in weather and viewpoint variation, yet struggles to achieve competitive performance in the presence of illumination variation. MixVPR shows top-end performance across all dataset variations, with itself, Cosplace and EigenPlaces achieving the best performance on \casMX{} by a wide margin. At $L=1$, these 3 models achieve close to AUC = 1 on \casMX{} and \luzMX{}, revealing their tolerance to illumination and weather differences. However, Cosplace and EigenPlaces degrade at a larger rate on \luzMX{} and \guaMX{}, so MixVPR may be better equipped to handle motion blur.
FloppyNet achieves competitive performance on \guaMX{}, yet in the presence of weather or illumination variance, it struggles to compete with the other methods. P-NetVLAD, ORB-VLAD and SAD stand out for their intolerance to these mixed conditions. P-NetVLAD appears to equally suffer across each dataset, starting close to 0.8 AUC and dropping to 0.2. Although ORB-VLAD and SAD showed some tolerance in the blur-only traverses at sharp and small blur intensity, they suffer in mixed conditions in any case and are incapable of VPR.

\subsection{Deblurring Query Images}

\begin{figure*}[!t]
\centering
\includegraphics[width=\textwidth]{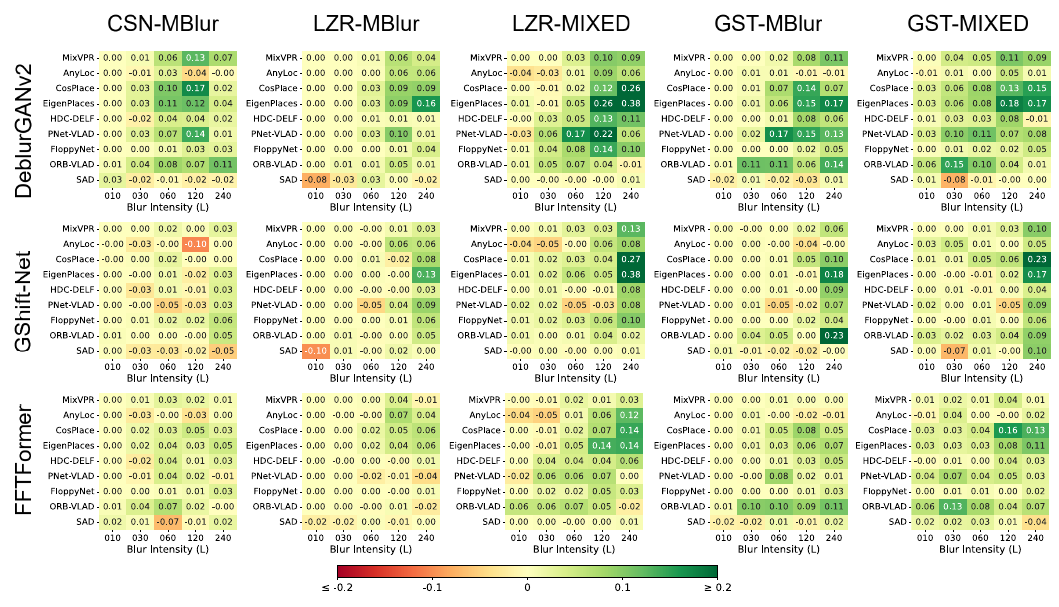}
\caption{Heatmaps showing the \textbf{difference} in AUC performance after deblurring the other datasets (columns) with each deblurring method (rows), with different VPR method and blur intensity combinations.}
\label{fig:heatmaps}
\vspace*{-1em}
\end{figure*}

We examine the impact of deblurring across various blur intensities to assess its effect on VPR accuracy. For \casMX{}, Table IV presents full results, including the combined average deblurring and query extraction times, while Fig. \ref{fig:heatmaps} shows performance differences for the other datasets. Deblurring effectiveness depends on the method, VPR model, and blur intensity, with greater improvements at severe blur levels. Overall, VPR models resilient to appearance variations achieve greater average AUC and lower AUC standard deviation across blur intensities due to enhanced image restoration.


Generally, CosPlace and EigenPlaces respond positively and consistently with deblurring. They use global image descriptors, making them highly scalable and benefiting from deblurring methods that restore spatial context across the image. MixVPR adopts a relatively simple approach too, using a holistic feature aggregation technique, and benefits 
in most cases with deblurring. Unexpectedly, the effectiveness of all three deblurring methods on AnyLoc performance was limited, with small improvements in \casMX{}, \luzMX{} and \luzmb{}, reinforcing the fact that it may already be inherently more tolerant to motion blur. Likewise, FloppyNet's strong performance in motion blur-only datasets means it benefits little from deblurring in those cases, however, in mixed conditions, it shows small improvements, particularly with DeblurGANv2. Deblurring provides some improvement for HDC-DELF under severe motion blur, but it is often minor and less impactful than for methods that rely exclusively on sharp, global image features. Both ORB-VLAD and P-NetVLAD show tangible improvement when combined with DeblurGANv2 across all datasets, highlighting their intolerance for motion blur, especially with the \guaMX{} dataset, which degraded all model performance significantly. Regardless of deblurring, SAD is incapable of effective VPR unless the images are sharp and do not contain complex variations.

DeblurGANv2 proves to be the most effective and robust choice of the three when applied to VPR, providing larger and more consistent improvements in VPR performance. Its strong performance can be attributed to its emphasis on multi-scale deblurring. Its use of a feature pyramid network enables deblurring across broad spatial resolutions, which may support larger and more complex blur patterns.
Unlike the other methods, GShift-Net is developed for video deblurring and implicitly aggregates temporal and spatial information in its shift blocks. By shifting feature groups across frames, it increases the receptive field. This method performs relatively poorer than the other three, however, interestingly, we can see that GShift-Net accounts for some of the best scores with blur level 240 by a large margin, such as when combined with Eigenplaces on \casMX{}, we see a gain of 0.152. This may indicate that adjacent frames could provide valuable information when context is lost under extreme blur conditions. Future work could explore the importance of temporal information under varying degrees of image degradation.
The impacts of FFTFormer resemble DeblurGANv2, however, it does not provide significant gains. More often than not, it still shows measurable improvement over no deblurring, and even when performance drops, it is minimal. These results are unexpected as FFTFormer achieves greater performance than DeblurGANv2 on 'GoPro', 'RealBlur' \cite{rim_2020_ECCV} and 'HIDE' evaluation. Their differences in training data may align more or less with the variations present in our benchmark. Notably, DeblurGANv2 and GShift-Net were further trained on the DVD \cite{Su_2017_CVPR} dataset, providing additional videos captured at 240 fps, and DeblurGANv2 on the NFS video dataset \cite{galoogahi2017need}. FFTFormer, however, focuses on RealBlur and HIDE datasets, highlighting the need for careful consideration of the domain when building deblurring VPR or SLAM systems.

\begin{table*}[!htp]
\centering
\scriptsize
\caption{Comparison of inference efficiency and AUC for different deblurring strategies on shuffled mixed-condition datasets, measured with the PyJoules library on an AMD EPYC 7452 CPU and Nvidia A100 GPU.}
\begin{tabularx}{\textwidth}{
  >{\centering\arraybackslash}p{4cm}   
  p{1.7cm}                          
  >{\centering\arraybackslash}X     
  >{\centering\arraybackslash}X     
  >{\centering\arraybackslash}X      
  >{\centering\arraybackslash}p{1.7cm}     
  >{\centering\arraybackslash}X      
  >{\centering\arraybackslash}X     
  >{\centering\arraybackslash}X     
  >{\centering\arraybackslash}p{1.7cm}      
  >{\centering\arraybackslash}X     
}
\toprule
\textbf{Setup} & \textbf{Method}
  & \multicolumn{4}{c}{\textbf{Timing (s)}}
  & \multicolumn{4}{c}{\textbf{Energy (kJ)}}
  & \textbf{AUC} \\
\cmidrule(lr){3-6}\cmidrule(lr){7-10}
 & & \textbf{Extract} & \textbf{Detect} & \textbf{Deblur} & \textbf{Avg/Query (ms)}
   & \textbf{Extract} & \textbf{Detect} & \textbf{Deblur} & \textbf{Avg/Query (J)}
   & \\
\midrule

\multirow{3}{*}{DeblurGANv2-CosPlace-\casMX{}}
 & No Deblur
   & 1.39 & --   & --   & 3.76
   & 0.41 & --   & --   & 1.12
   & 0.93 \\
 & All Deblur
   & 1.40 & --   & 12.80 & 38.48
   & 0.42 & --   & 2.80 & 8.75
   & 0.97 \\
 & Detect+Deblur
   & 1.39 & 0.71 & 7.76 & 26.72
   & 0.42 & 0.10 & 1.73 & 6.10
   & 0.97 \\
\midrule

\multirow{3}{*}{DeblurGANv2-EigenPlaces-\guaMX{}}
 & No Deblur
   & 1.28 & --   & --   & 3.78
   & 0.40 & --   & --   & 1.18
   & 0.69 \\
 & All Deblur
   & 1.28 & --   & 11.69 & 38.26
   & 0.38 & --   & 2.57 & 8.70
   & 0.75 \\
 & Detect+Deblur
   & 1.29 & 0.65 & 9.41 & 33.48
   & 0.40 & 0.094 & 2.00 & 7.36
   & 0.75 \\
\midrule

\multirow{3}{*}{DeblurGANv2-EigenPlaces-\luzMX{}}
 & No Deblur
   & 1.34 & --   & --   & 3.76
   & 0.41 & --   & --   & 1.15
   & 0.93 \\
 & All Deblur
   & 1.34 & --   & 12.42 & 38.65
   & 0.41 & --   & 2.74 & 8.85
   & 0.95 \\
 & Detect+Deblur
   & 1.35 & 0.72 & 6.42 & 23.85
   & 0.41 & 0.10 & 1.39 & 5.34
   & 0.95 \\

\bottomrule
\end{tabularx}
\vspace*{-1em}
\label{tab:adptv_Deblur}
\end{table*}

\subsection{Adaptive Deblurring}

Results have shown that certain models such as Cosplace and Eigenplaces exhibit top VPR performance on the benchmark, being more robust to scene variations and responding well to deblurring. Regardless, the performance gains are often small or even degraded with small motion blur intensity, motivating an adaptive deblurring approach which may achieve improvement in computational cost with minimal change in performance. 


Table \ref{tab:adptv_Deblur} compares several significant adaptive deblurring scenarios using the best method, DeblurGANv2, which yielded larger gains across blur levels (see \fig{fig:heatmaps}). 
We measured inference time and energy consumption using PyJoules on a single AMD EPYC 7452 CPU and Nvidia A100 GPU. To avoid null values in short per-image measurements - a limitation of the profiling software, we measured the total cost of inference in each stage separately and aggregated them, reflecting the cumulative inference overhead while ensuring consistent and reliable values across all scenarios.
Based on prior results, adaptive strategies are justified only in scenarios with frequent severe blur. To represent these scenarios, the shuffled datasets \casMX{} and \luzMX{} include roughly half sharp ($L=1$) images, with the remainder spanning blur levels $L=60$ to $L=240$. The \guaMX{} shuffled dataset, which exhibited relatively less improvement, contains $\approx 20\%$ sharp, the rest at blur level $L=120$ and $L=240$. The same adaptive method can be applied to specific cases with other VPR or deblur methods, such as FFTFormer and Anyloc in \luzMX{}, which had degradation with small motion blur.
In these shuffled mixed-condition scenarios, an all-deblur strategy yields an AUC gain between 0.02 and 0.06. Given that FFTFormer and GShift-Net did not perform as well in previous analyses, they would struggle more in mixed datasets, making them unsuitable for real adaptive deblurring regardless of their computational overhead. Having used the largest backbone, Inception-ResNet-v2, for DeblurGANv2, both the energy consumption and total time significantly increase (processing $\sim$25 fps). In comparison, using a prior detection stage achieves the same performance gain with less additional overhead, as seen in the shuffled \casMX{} and \luzMX{} scenarios. The proportion of sharp images affects the trade-off between computational cost and performance, with sharp images avoiding redundant and costly deblurring operations, while blurred images, though requiring additional time and energy expenditure, enable potential gains in recognition accuracy. Adaptive deblurring is shown to be beneficial over direct deblurring when the VPR model tolerates complex scene variations and when severe, frequent blur allows for significant restoration improvements. In these cases, since most of the overhead comes from deblurring, an adaptive strategy offers substantial benefits.

\subsection{Practical Implications}

While these systems run sufficiently fast on GPUs, deploying them on resource-constrained, motion-blur-prone devices like UAVs is challenging. \cite{drones5040121} used a Raspberry Pi v2.1 camera for non-learning-based VSLAM, which was limited to 20 fps for successful tracking, even with offline detection and deblurring. Traditional deblurring methods (e.g., Wiener, Lucy-Richardson) are also iterative and slow for large images, limiting real-time use. Consequently, one could explore techniques like quantization in deblurring pipelines \cite{Chiang_2020_CVPR_Workshops}, developing blur-robust VPR methods that bypass explicit deblurring, or employing spiking neural networks \cite{snn} with event-based cameras or neuromorphic hardware - a promising avenue for improving efficiency.


\section{Conclusions}
\label{sec:conclusions}


Our benchmark enables comprehensive VPR analysis under motion blur. In certain configurations, combining VPR methods with deblurring improves matching performance and robustness, though real-time adaptive deblurring on resource-limited devices remains challenging despite improvements in efficiency when blur is significant.
Future work could extend the blur model to other domains, or explore end-to-end sequence-based pipelines that make use of temporal information \cite{Li_2023_CVPR}, \cite{consecutive}. In extreme conditions (e.g., adverse weather), generic deblurring may struggle. Exploring domain-adaptive deblurring or multi-modal sensing using event-based cameras, LiDAR or IMU data could improve accuracy, particularly in high-speed or low-light scenarios.

%

\bibliographystyle{bibliography/IEEEtran}
\bibliography{bibliography/IEEEabrv,bibliography/references}

\end{document}